\documentclass[twocolumn, 10pt]{IEEEtran}
\usepackage{graphicx}
\usepackage{amssymb}
\usepackage{amsmath}
\usepackage{mathtools}
\usepackage{dsfont}
\usepackage{cite}
\usepackage{stfloats}
\usepackage{subfigure}
\usepackage{psfrag}
\usepackage[mathscr]{euscript}
\usepackage{algorithm, algorithmic}
\usepackage{acronym}  % make an acronym
\usepackage{booktabs}
\usepackage{bbm}

\usepackage{multirow}

\usepackage{multicol}

\usepackage{xcolor}
\usepackage{ulem}
\usepackage{float}
\usepackage{balance}

\acrodef{CCDF}{complementary cumulative distribution function}
\acrodef{CF}{characteristic function}
\acrodef{PPP}{Poisson point processe}
\acrodef{RV}{random variable}
\acrodef{i.i.d.}{independent and identically distributed}
\acrodef{PDF}{probability distribution function}
\acrodef{CDF}{cumulative distribution function}
\acrodef{ch.f.}{characteristic function}
\acrodef{AWGN}{additive white Gaussian noise}
\acrodef{SNR}{signal-to-noise ratio}
\acrodef{LRT}{likelihood ratio test}
\acrodef{DRT}{distance ratio test}
\acrodef{GLRT}{generalized likelihood ratio test}
\acrodef{CRLB}{Cram\'{e}r-Rao lower bound}
\acrodef{CRB}{Cram\'{e}r-Rao bound}
\acrodef{ZZLB}{Ziv-Zakai lower bound}
\acrodef{ZZB}{Ziv-Zakai bound}
\acrodef{LOS}{line-of-sight}
\acrodef{ToF}{time-of-flight}
\acrodef{NLOS}{non-line-of-sight}
\acrodef{GDOP}{geometric dilution of precision}
\acrodef{GPS}{Global Positioning System}
\acrodef{FIM}{Fisher information matrix}
\acrodef{PEB}{position error bound}
\acrodef{SPEB}{squared position error bound}
\acrodef{TOA}{time-of-arrival}
\acrodef{TOF}{time-of-flight}
\acrodef{WSN}{wireless sensor network}
\acrodef{MAC}{medium access control}
\acrodef{RSS}{received signal strength}
\acrodef{WAF}{wall attenuation factor}
\acrodef{TDOA}{time difference-of-arrival}
\acrodef{RF}{radiofrequency}
\acrodef{RTT}{round-trip time}
\acrodef{AOA}{angle-of-arrival}
\acrodef{MF}{matched filter}
\acrodef{ED}{energy detector}
\acrodef{ML}{maximum likelihood}
\acrodef{MSE}{mean-square error}
\acrodef{RMSE}{root-mean-square error}
\acrodef{LEO}{localization error outage}
\acrodef{ppm}{part-per-million}
\acrodef{ACK}{acknowledge}
\acrodef{UWB}{Ultrawide bandwidth}
\acrodef{TNR}{threshold-to-noise ratio}
\acrodef{LS}{least squares}
\acrodef{IR-UWB}{impulse radio UWB}
\acrodef{FCC}{Federal Communications Commission}
\acrodef{TH}{time-hopping}
\acrodef{PPM}{pulse position modulation}
\acrodef{MUI}{multi-user interference}
\acrodef{PDP}{power delay profile}
\acrodef{BPZF}{band-pass zonal filter}
\acrodef{SIR}{signal-to-interference ratio}
\acrodef{SINR}{signal-to-interference-plus-noise ratio}
\acrodef{RFID}{radio frequency identification}
\acrodef{WPAN}{wireless personal area network}
\acrodef{WWB}{Weiss-Weinstein bound}
\acrodef{DP}{direct path}
\acrodef{MF}{matched filter}
\acrodef{MMSE}{minimum-mean-square-error}
\acrodef{SBS}{serial backward search}
\acrodef{SBSMC}{serial backward search for multiple clusters}
\acrodef{NBI}{narrowband interference}
\acrodef{WBI}{wideband interference}
\acrodef{INR}{interference-to-noise ratio}
\acrodef{CR}{channel response}
\acrodef{CIR}{channel impulse response}
\acrodef{CR}{channel  response}
\acrodef{RADAR}{radar}
\acrodef{MUR}{Multistatic radar}
\acrodef{JBSF}{jump back and search forward}
\acrodef{HDSA}{high-definition situation-aware}
\acrodef{RRC}{root raised cosine}
\acrodef{ST}{simple thresholding}
\acrodef{BTB}{Bellini-Tartara bound}
\acrodef{P-Max}{$P$-Max}  %suggestion, use with \acl{P-Max}
\acrodef{MIMO}{multiple-input multiple-output}
\acrodef{MAP}{maximum a posteriori}
\acrodef{FG}{factor graph}
\acrodef{OP}{outage probability}
\acrodef{WED}{wall extra delay}
\acrodef{RMS}{root mean square}
\acrodef{SPAWN}{sum-product algorithm over a wireless network}
\acrodef{MDD}{minimum distance distribution}
\acrodef{MAP}{maximum a posteriori probability}
\acrodef{SAP}{small cell access point}
\acrodef{UE}{user equipment}
\acrodef{MBS}{macro cell base station}
\acrodef{UER}{\ac{UE} Relay}
\acrodef{D2D}{device-to-device}
\acrodef{MBS}{macro base station}
\acrodef{CSI}{channel state information}
\acrodef{OGR}{outage guard region}
\acrodef{FUR}{feasible UER region}
\acrodef{EHR}{energy harvesting region}
\acrodef{EH}{energy harvesting}
\acrodef{D2D-EHSN}{D2D communication provided \ac{EH} small cell network}
\acrodef{D2D-EHHN}{D2D communication provided \ac{EH} heterogeneous network}
\acrodef{3GPP}{3rd Generation Partnership Project}
\acrodef{BS}{base station}
\acrodef{DF}{decode and forward}
\acrodef{CCDF}{complementary cumulative distribution function}
\acrodef{ZF}{zero forcing}
\acrodef{RZF}{regularized zero forcing}
\acrodef{WLLN}{weak law of large number}
\acrodef{SLLN}{strong law of large numbers}
\acrodef{TDD}{Time-division duplex}
\acrodef{EE}{energy efficiency} 
\acrodef{HetNet}{heterogeneous network} 
\acrodef{SCP}{Single Cell Processing}
\acrodef{CBF}{Coordinated Beamforming}
\usepackage{color}
\usepackage{dsfont}
\usepackage{bbm}

\DeclareMathAlphabet{\mathsf}{OML}{cmbr}{m}{it}

\newcommand{\bd}{\begin{description}}
\newcommand{\ed}{\end{description}}
\newcommand{\be}{\begin{enumerate}}
\newcommand{\ee}{\end{enumerate}}
\newcommand{\bi}{\begin{itemize}}
\newcommand{\ei}{\end{itemize}}
\newcommand{\bl}{\begin{list}}
\newcommand{\el}{\end{list}}
\newcommand{\bt}{\begin{tabbing}}
\newcommand{\et}{\end{tabbing}}

\newcommand{\paperTitle}{Edge Intelligence Over the Air: Two Faces of Interference in Federated Learning}

\newcommand{\ota}{over-the-air}

\begin{document}

\title{\paperTitle}

\author{
    Zihan~Chen, \textit{Member, IEEE},
    Howard H. Yang, \textit{Member, IEEE},
    and Tony Q. S. Quek, \textit{Fellow, IEEE}

\thanks{This paper is supported in part by the National Research Foundation, Singapore and Infocomm Media Development Authority under its Future Communications Research \& Development Programme, in part by the National Natural Science Foundation of China under Grant 62271513, in part by the Zhejiang Provincial Natural Science Foundation of China under Grant LGJ22F010001, and in part by the Zhejiang - Singapore Innovation and AI Joint Research Lab. (\textit{Corresponding authors: T. Q. S. Quek, H. H. Yang})}

\thanks{Z. Chen and T. Q. S. Quek are with the Singapore University of Technology and Design, Singapore 487372. T. Q. S. Quek is also with the Department of Electronic Engineering, Kyung Hee University, Yongin 17104, South Korea (e-mail: \{zihan\_chen, tonyquek\}@sutd.edu.sg).}

\thanks{H. H. Yang is with the Zhejiang University/University of Illinois Urbana-Champaign Institute, Zhejiang University, Haining 314400, China, the College of Information Science and Electronic Engineering, Zhejiang University, Hangzhou 310007, China, and the Department of Electrical and Computer Engineering, University of Illinois Urbana-Champaign, Champaign, IL 61820, USA (email: haoyang@intl.zju.edu.cn).}

% \thanks{T. Q. S. Quek is with the Singapore University of Technology and Design, Singapore 487372, and also with the Department of Electronic Engineering, Kyung Hee University, Yongin 17104, South Korea (e-mail: tonyquek@sutd.edu.sg). (\textit{Corresponding authors: T. Q. S. Quek, H. H. Yang})}

}
\maketitle
\acresetall
\thispagestyle{empty}
%%---------------------------------------------------------------------------%
%%                           abstract and key words                          %
%%---------------------------------------------------------------------------%
\begin{abstract}
Federated edge learning is envisioned as the bedrock of enabling intelligence in next-generation wireless networks, but the limited spectral resources often constrain its scalability. 
In light of this challenge, a line of recent research suggested integrating analog over-the-air computations into federated edge learning systems, to exploit the superposition property of electromagnetic waves for fast aggregation of intermediate parameters and achieve (almost) unlimited scalability. 
Over-the-air computations also benefit the system in other aspects, such as low hardware cost, reduced access latency, and enhanced privacy protection. 
Despite these advantages, the interference introduced by wireless communications also influences various aspects of the model training process, while its importance is not well recognized yet. This article provides a comprehensive overview of the positive and negative effects of interference on over-the-air computation-based edge learning systems. The potential open issues and research trends are also discussed.
\end{abstract}
\acresetall

%%---------------------------------------------------------------------------%
%%                           Sec: Introduction                               %
%%---------------------------------------------------------------------------%
\section{Introduction}\label{sec:intro}
% With the rapid advancement of Internet-of-Things (IoT) and edge computing, sheer volumes of data are generated and collected by numerous edge devices. 
% The increasing concerns about the data privacy in network management and the asymmetry property of network connection, make it not practical to conduct centralized  training via transmitting user data to the central server. 
The rapid development of Internet-of-Things (IoT), artificial intelligence (AI), and edge computing is forging ahead a variety of emerging applications, ranging from autonomous driving and smart healthcare to intelligent network management, that significantly improve the quality of life of humans. 
These applications rely on advanced machine learning (ML) algorithms that train statistical models based on the sheer volumes of data collected from various edge nodes, enabling devices to make decisions in accordance with local events. 
However, traditional ML approaches require users to gather their data in a central server for model training. 
This reveals oppressive privacy and security consequences in which end-users' sensitive information may be shared with and/or exploited by other entities, hence intensifying the demand for new solutions.
With the advances in communication techniques and the improvements in the computation capabilities of edge devices, distributed learning, especially federated learning (FL), provides a promising solution for widely dispersed  edge  devices to collaboratively train a global ML model without sharing the private data \cite{MaMMooRam:17AISTATS}.
Recently, FL over wireless network, which involves deploying ML models at the network edge, i.e., \textit{federated edge learning}, has gained significant attention~\cite{ParSamBen:19,yang_tcom_fl_scheduling}. 
% via deploying ML models at the edge of the network, i.e., federated edge learning, gains growing attentions

A typical federated edge learning system consists of an edge server and multiple user equipments (UEs), in which a global model is shared amongst the entities.
In each iteration, a portion of UEs is selected to conduct on-device model training based on their private dataset. 
Then, the server collates the local gradients to improve the global model and broadcasts the updated result back to a new subset of selected UEs for the next round of local computations.
Such interactions between the edge server and UEs would repeat until the model converges.
Generally, it takes hundreds, even thousands, of communication rounds to reach convergence.
In view of the hefty communication overhead incurred by the iterative exchange of intermediate parameters, which becomes particularly cumbersome in the digital communication-based federated edge learning system where every UE needs an orthogonal sub-channel to upload its local update, a new design that adopts \textit{analog over-the-air computations} for model aggregation is proposed to address such a communication bottleneck \cite{ding2020ota,zhu2021otanetwork,YanCheQue:21JSTSP}. 
The key idea of analog over-the-air computations is that UEs modulate their local gradients onto a set of common waveforms and simultaneously send out the analog signals to the edge server, exploiting the superposition property of multi-access channels to realize fast aggregation of the gradients \cite{guo2021ota}. 
Incorporating this technique into the federated edge learning system has the following benefits: 
\begin{itemize}
    \item \textit{High spectral efficiency:} Compared with the conventional digital communication-based FL that needs to select a portion of the UEs for parameter updating  in each communication round, analog \ota{} computing dramatically boosts up the spectral utilization by allowing all the UEs to simultaneously access the spectrum and upload their parameters to the edge server, irrespective of the number of UEs present in the system. As a result, the bandwidth no longer constrains the number of participating UEs in the system, expediting the large-scale deployment of edge learning systems \cite{zhu2021otanetwork}.
    \item \textit{Low communication cost:} Analog \ota{} computations can be achieved by elementary communication techniques such as amplitude modulation. Additionally, the UEs are also disburdened with estimating the instantaneous channel state information before each global transmission. Moreover, as opposed to digital communication-based FL, an increase in the number of UEs in the network can improve the system's energy efficiency, thereby enabling the UEs to reduce their transmit power \cite{YanCheQue:21JSTSP}. These salient features make it possible to establish an intelligent edge system over massively distributed UEs with low-cost communication modules for cognitive network management. 
    \item \textit{Improved privacy protection:} 
    The channel fading and interference noise induced by analog \ota{} computation impose a random perturbation to each UE's uploaded parameter. As a result, even if there are eavesdroppers in the network, no original information can be recovered from the aggregated gradients. Such an enhancement to privacy protection in network management is particularly relevant in privacy-aware edge learning systems \cite{elgabli2021harnessingadmm,liu2020privacyfree}.
    \item \textit{Reduced access latency:} By virtue of analog over-the-air computing, the local updates would be automatically aggregated at the edge server's output. This inherent integration of communication and computation significantly reduces the access, as well as processing latency, since the system no longer needs to go through the encoding (resp. decoding) and modulation (resp. demodulation) processes to obtain the individual gradients of each UE before adding them up \cite{ding2020ota, zhu2019broadband}.
\end{itemize}

\begin{figure*}[t!]
  \centering{\includegraphics[width=0.95\linewidth]{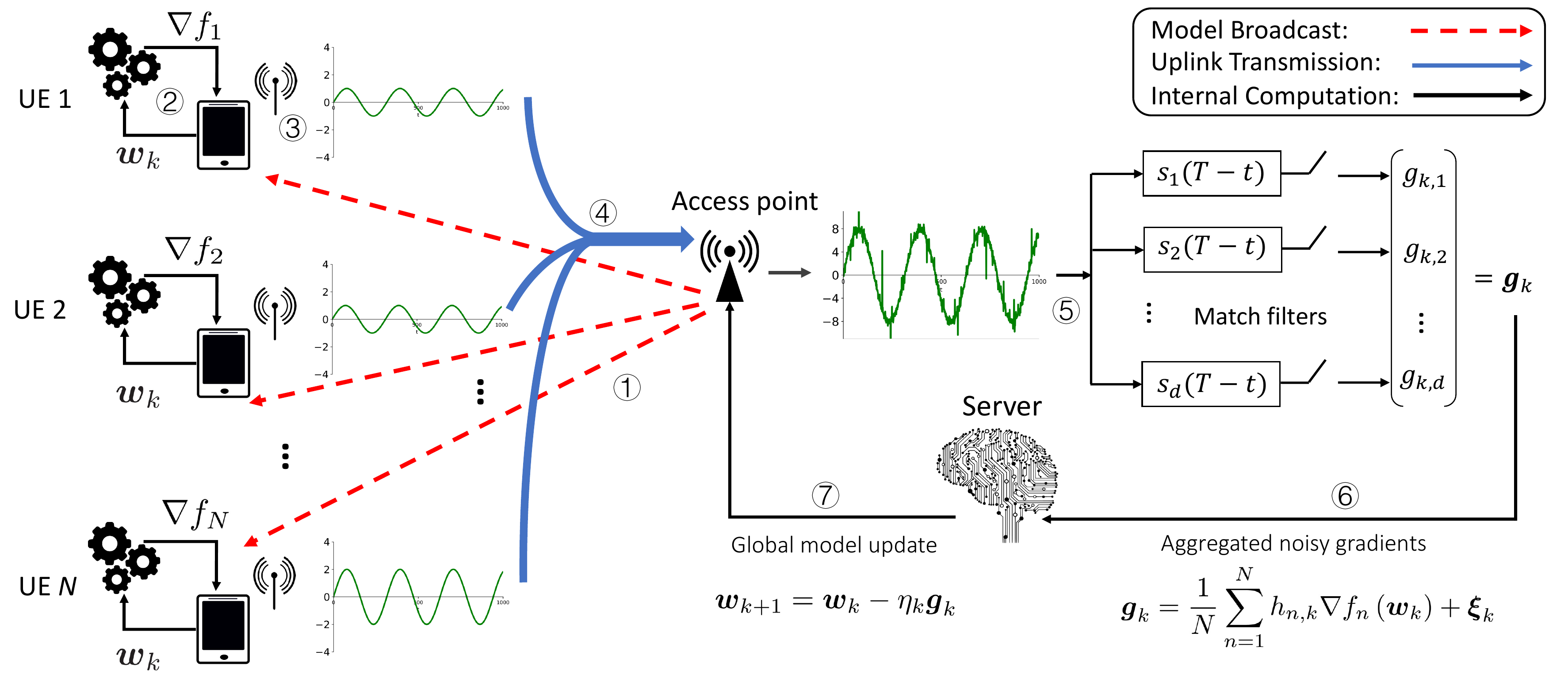}}
  \caption{An overview of the analog over-the-air federated edge learning system. Local gradients are uploaded using amplitude modulations. Aggregation is  then performed automatically in the air, where a heavy-tailed noise with tail index $\alpha=1.5$ is added to the aggregated global gradient signal. Steps of the framework in a typical communication round are numbered accordingly.}
  \label{fig:ota_system_flow}
\end{figure*}

Despite the above advantages, the uncoded analog transmissions also expose UEs' gradients to channel fading and interference noise, which can afflict the received signal quality, leading to a degradation in the system performance. %inflict
That said, in the context of ML, noise does not merely have a negative effect. 
Indeed, through appropriate signal processing techniques, we can turn the interference noise from a setback into an asset for federated edge learning systems \cite{YanCheQue:21JSTSP,zhang2022turningpca}.
This two-sided effect of interference results in several important operational changes in analog \ota{}-based FL, which is not widely known.
The central goal of the present article is, therefore, to summarize the key research findings related to the interference in analog \ota{} computing and their implications in the design of federated edge learning systems, and more importantly, this article seeks to provide insights for the new design of the analog over-the-air federated edge learning system via taming and repurposing the channel interference.

\section{Analog Over-the-Air Machine Learning}
This section gives a brief overview of FL under analog \ota{} computations, also coined as analog \ota{} machine/edge learning. 
Note that the main impact of analog \ota{} computing is on alleviating burdens in the uplink transmission, i.e., from UEs to the edge server, which is the well-known ``communication bottleneck" of the system \cite{shi2021otaisit}. 
% In contrast to the edge learning systems based on the digital communication techniques, the main idea of the analog \ota{} machine learning is that UEs transmit their locally computed gradients/model updates in the form of uncoded analog signals to the edge server.
% , in which the amplitude modulation is considered in the uplink communications. 
% Note that we only consider gradients or model updates in uplink communication for simplicity. The analog \ota{}-computation remains applicable when full local models are transmitted in the uplink connection of edge learning systems, as the aggregation over the  model weights enjoys the same computation complexity while compared with the gradient aggregation.
\subsection{General Framework}
The training process of the analog \ota{} edge learning is detailed as follows (see Fig. \ref{fig:ota_system_flow} for an overview). Let us consider an edge learning system in which $N$ UEs collaboratively train a global ML model under the coordination of an edge server. 
To be more concrete, let $f_i(\cdot):\mathbb{R}^d \rightarrow \mathbb{R}$ be the empirical loss function constructed via the local dataset of UE $i$. The goal of all the entities in the system is to solve the following optimization problem:
\begin{equation}\label{eq:min_prob}
    \underset{{ \boldsymbol{w} \in \mathbb{R}^d }}{ \mathrm{min} } ~~ f(\boldsymbol{w}) = \frac{1}{N} \sum_{ n=1 }^N f_n( \boldsymbol{w} ),
\end{equation}
where $\boldsymbol{w} \in \mathbb{R}^d$ denotes the model parameters and $N$ represents the total number of UEs.
The learning task in Eq.~\eqref{eq:min_prob} is accomplished by means of gradient descent (GD) training method and analog \ota{} computing (See illustration in Fig.~\ref{fig:ota_system_flow} with steps numbered accordingly).
Specifically, in a typical communication round $k$, the model training contains the following steps: (1) The latest global model is distributed by the edge server to each UE $i$. (2) Each UE $i$ computes its local gradients $\nabla f_i$ and then (3) modulates the gradient on a common set of orthonormal waveforms, in which the amplitude of each waveform is tuned in accordance with the value of a distinct element of the gradient \cite{ding2020ota}.
All UEs transmit their modulated waveforms over the air in a synchronized manner and the signals arrive at the edge server simultaneously \cite{zhu2021otanetwork,guo2021ota}\footnote{The precise synchronization can be achieved by appropriately setting the cyclic prefix (CP) in an orthogonal frequency division multiplexing (OFDM) symbol or leveraging other state-of-the-art protocols used in cellular systems.}.
Benefiting from the superposition property of multiple access channels, (4) all the signals would be automatically aggregated in the air.
(5) After passing the received signal to a bank of match filters that are tuned according to the waveforms $s_i(T-t)$, the edge server can obtain the aggregated global gradients from the output.
Owing to the effects of the channel fading and interference on analog signals, (6) the aggregated global gradients obtained by the edge server in $k$-th communication round can be expressed as follows:
\begin{equation}\label{eq:ota_agg}
\boldsymbol{g}_{k}=\frac{1}{N} \sum_{n=1}^{N} h_{n, k} \nabla f_{n}\left(\boldsymbol{w}_{k}\right)+\boldsymbol{\xi}_{k},
\end{equation}
where the $h_{n, k}$  denotes the channel fading for UE $n$ and $\boldsymbol{\xi}_{k}$ is a vector representing the interference noise~\cite{YanCheQue:21JSTSP}.

Aided with the aggregated (noisy) gradients, (7) the edge server updates the global model by gradient descent, and then broadcasts the new model to all UEs for the next round of computations. 
Overall, as a coordinator in the system, the edge server is responsible for aggregating and processing the analog signals, performing the global model update, and broadcasting the new model to the UEs, which plays an important role in spatial-temporal connection during training.

\subsection{A Closer Look at Interference }
The constructive property of electromagnetic waves often leads to the effects of strong repulsion in the interference noise. 
A direct consequence of this effect is the spikes in the edge server's received signal, as illustrated in Fig.~1. 
From the perspective of statistics, such a phenomenon indicates that the interference distribution is heavy-tailed. 
And it has been widely accepted that a symmetric $\alpha$-stable distribution can appropriately capture the heavy-tailed characteristic of interference and hence is often used for modeling interference in wireless networks. 
The parameter $\alpha$ is commonly known as the tail index and the detailed expression of the characteristic function can be found in \cite{YanCheQue:21JSTSP}.
% \sout{
% The formal definition of $\alpha$-stable distribution is given by the following:}

% \textbf{Definition:} \textit{The random variable $\xi$ is said to follow a symmetric $\alpha$-stable distribution if its characteristic function takes the following form: 
%     \begin{align}
%     \mathbb{E}\left[ e^{j \omega \xi } \right] = \exp( - \delta^\alpha \vert \omega \vert^\alpha )
%     \end{align}
%     where $\delta>0$ and $\alpha \in (0, 2]$. The parameters $\delta$ and $\alpha$ are commonly known as the scale parameter and tail index, respectively.  
% }

Notably, $\alpha$-stable distributions generally have no analytical expressions of the probability density function (PDF), except for two special cases, namely, if $\alpha = 1$, the distribution reduces to Cauchy, and when $\alpha = 2$, it reduces to Gaussian.
Moreover, a random variable that follows $\alpha$-stable distribution has finite moments only up to the $\alpha$ order, beyond which the moment is unbounded~\cite{simsekli2019tailindex}. 

% {\color{red}{Fig.~\ref{fig:heavy-tail}(a) depicts the PDF of a symmetric $\alpha$-stable random variable under different tail indices. 
% We can see that the smaller the $\alpha$, the heavier the tail in PDF. }}
In general, an $\alpha$-stable random variable with a smaller value of $\alpha$  ensues a heavier tail in the distribution.
Correspondingly, the interference noise has a higher chance of generating repulsive responses: As illustrated by Fig.~\ref{fig:heavy-tail}, interference noise with a smaller tail index (i.e. $\alpha = 1.2$) results in more huge spikes in the analog signal. 
Note that we shall adopt different values for $\alpha$ across this work to demonstrate the effects of interference.
According to (2), these spikes severely deviate from the received signal values, which in turn distorts the aggregated gradients.
To this end, the following discussions are centered around the effects of the interference characteristics on the various aspects of system performance~\cite{YanCheQue:21JSTSP}. 

\section{The Negative Side of Interference} \label{sec:bad_face}
% In the analog \ota{} computations, the analog signals suffers distortions due to the random channel fading and interference in the wireless network. The consequent noisy aggregated global gradients would then result in performance degradation of the updated global model. 
% A natural question may arise as: \textit{Does analog \ota{} edge learning algorithms converge with the noisy gradients?} 
% Thus, this section would mainly address such performance degradation on these two aspects: convergence and performance stability.
Interference noise inflicts degradation on wireless transmissions. 
Analog \ota{} computing is not immune from that. 
In the context of analog over-the-air FL, the main detriments stem from interference are the following: 

\begin{figure}[t!]
  \centering
  \includegraphics[width=0.95\columnwidth]{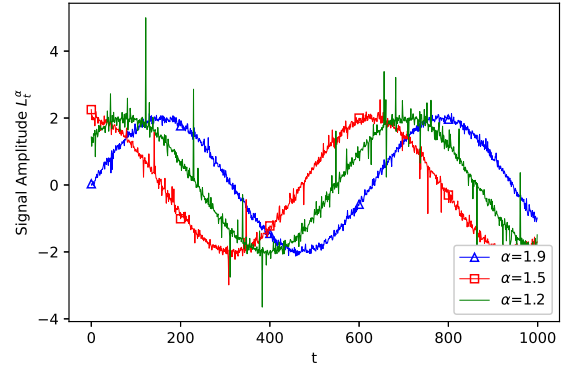}  % used in R2
  \caption{ An illustrative example of the effects of heavy-tailed noise imposed on three different sinusoid signals, in which smaller $\alpha$ leads to more huge spikes (i.e., higher amplitude) on the analog signal, while a larger $\alpha$ exhibit fewer impacts.}
  \label{fig:heavy-tail}
\end{figure}

%A depiction of $\alpha$-stable random variables: ($a$) plots the probability density function under different values of the tail index and ($b$) illustrates the effects of heavy-tailed noise imposed on three different sinusoid signals.

\subsection{Slower Convergence Rate}
% According to the nature of heavy-tailed distribution, it is indispensable to consider the possibility of the strong impulsive noise from electromagnetic interference (see Fig. \ref{fig:2b}).
% In some communication rounds, such unneglectable impulsive interference would significantly sheer the corresponding magnitude of the global aggregated noisy gradients (see \eqref{eq:ota_agg}) and lead to severe distortion. 
% In the GD-based learning process, such noisy global gradients may bring unexpected divergence for subsequent optimization process due to the unanticipated extreme value.
The nature of heavy-tailed distribution in electromagnetic interference makes it indispensable to consider the possibility of strong impulsive noise in the aggregated global gradients. 
In the GD-based learning process, the unanticipated extreme value in interference power can significantly distort the global gradients, leading to unexpected divergence for the subsequent optimization processes. 
As such, the first fundamental question is: \textit{Does analog \ota{} edge learning algorithms converge with the noisy gradients?} 

% As depicted in Fig. \ref{fig:2a}, the tail index $\alpha$ controls the heaviness of the tail in the probability density function. 
% Specifically, the value of $\alpha$ would affects the magnitude of the extreme large impulse on analog signals and the probability of the extreme value of interference.
% It is found that different index $\alpha$ would essentially influence the convergence rate from the empirical aspects. 

\begin{figure}[t!]
  \centering
    {\includegraphics[width=0.95\columnwidth]{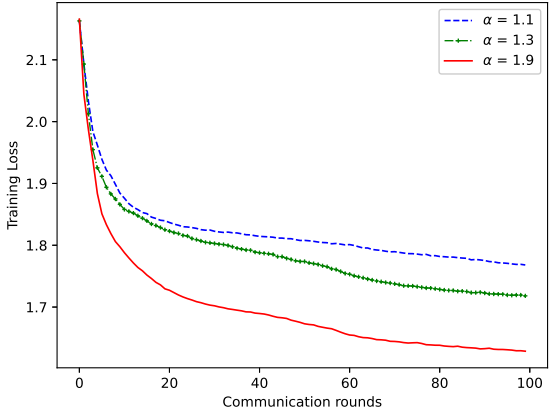}}
  \caption{ Simulation results of the training loss of a multi-layer-perceptron (MLP) on the MNIST dataset, with different tail index $\alpha$. }
  \label{fig:alpha_converge}
\end{figure}

Fig.~\ref{fig:alpha_converge} gives the experimental result that compares the convergence performance under different tail index $\alpha$, where $100$ UEs jointly train a neural network to accomplish an image classification task on MNIST dataset via the analog \ota{} edge learning system \cite{YanCheQue:21JSTSP}.
This figure plots the training loss as a function of communication rounds and it conveys a two-fold message: ($a$) GD-based model training is resilient to parameter distortion and hence relevant to the edge learning system and ($b$) the tail index $\alpha$ plays a pivotal role in convergence rate. 
Specifically, as the tail index decreases from $\alpha = 1.9$ (relatively light-tailed) to $\alpha = 1.1$ (relatively heavy-tailed), there is an orders-of-magnitude slowdown in the convergence rate.

In fact, if the global loss function $f(\boldsymbol{w})$ given in (1) possesses good structural properties (i.e., strong convexity and smoothness), it can be shown that the algorithm converges in the order of $\mathcal{O}(1/k^{\alpha-1})$, where $k$ is the number of global iterations. 
This mathematical expression also confirms that the heavy-tailedness of interference distribution is a dominating factor in the convergence rate of analog \ota{} machine learning, whereas the heavier the tail, the slower the algorithm converges. 

\subsection{Unstable Training Performance}
The heavy-tailed interference not just decelerates the convergence rate of FL, but more crucially, incurs severe disturbance to the model training process. 
% would also bring large fluctuations due to the potential strong impulse, along with the negative impacts on the convergence rate.
Specifically, during the optimization steps of the GD, large discrepancy and divergence would be caused by the aggregated gradients with the possibly unexpected extreme-value noise (see Fig. \ref{fig:heavy-tail} for an illustrative example of strong impulses in the interference). 
Hence, observing relatively large fluctuations during the training process would be inevitable.
Note that this can be readily implemented on the existing wireless infrastructure, e.g., let UEs send the parameters using the orthogonal frequency-division multiplexing (OFDM) modulation in which the (discrete) Fourier basis already provides a set of orthogonal “waveforms". A similar phenomenon can be found in \cite{guo2021ota}, where we can clearly see that the algorithm converges in the presence of wireless noise and the noise leads to the spikes in the convergence curve.

\begin{figure}[t!]
  \centering
    {\includegraphics[width=0.95\columnwidth]{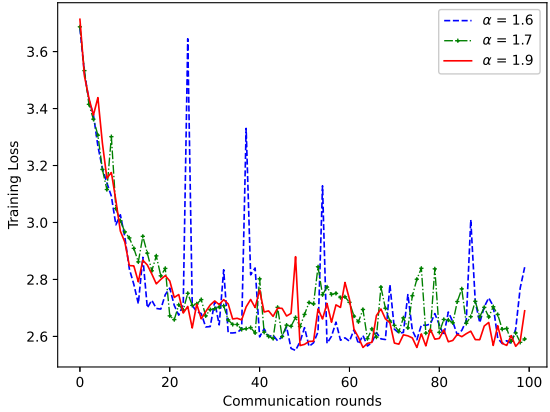}}
  \caption{ Simulation results of the training loss of LASSO problem solved via ADMM over synthetic data, with different tail index $\alpha$. }
  \label{fig:alpha_stability}
\end{figure}

Fig.~\ref{fig:alpha_stability} provides a pictorial demonstration of this phenomenon. Particularly, it considers an edge learning system in which 100 UEs jointly carry out a LASSO task based on distributed ADMM (\textit{i.e.}, alternating direction method of multipliers) and analog \ota{} computations. 
The figure clearly shows that decreasing the tail index $\alpha$, which increases the level of heavy-tailedness, results in more variations in the convergence curve. 
Besides, even in the regime of relatively light-tailed interference (namely, $\alpha = 1.9$), there is also unsteadiness associated with the convergence curve.
Part of the reason is that compared to GD-based algorithms, ADMM-based approaches are more sensitive to the variants in the intermediate parameters. 
This experiment demonstrates that the interference threatens the system's stability. 

\subsection{Inadequate Analytical Framework}
When it comes to the theoretical aspects, the conventional analysis framework relies on the existence of the gradients' second moments and may not be applicable when the tail index of the interference is small, i.e., $\alpha<2$, since in this case, the variance of the aggregated gradients given in (2) is infinite.
It thus entails the establishment of a new theoretical framework to quantify the convergence performance in analog \ota{} edge learning. 

In order to develop an analytical framework that is universally applicable to different tail indices of the interference, one shall opt for the $\alpha$-norm to quantify the difference in the model parameters upon each global updating step \cite{YanCheQue:21JSTSP}. As the expected value of aggregated gradients under this new distance metric can now be bounded, convergence analysis can be carried out via recursively bounding the variations in the global parameters and then conducting telescoping. 
It is also noteworthy that when $\alpha=2$, the $\alpha$-norm reduces to the conventional Euclidean norm, so as the convergence analysis. 

Nonetheless, there are only a few initial works on this research topic, where the current results heavily rely on assumptions that the objective function is strongly convex and smooth.
These presumptions have constrained the power of the analytical tools because many popular ML models, e.g., neural networks, are highly non-convex. 

% , which are confined in the constrained set of machine learning models. 

% In this case, $\alpha$ norm shall be opted for an universal analysis framework. 
% (Note that the $\alpha$-norm-based analysis framework also works when $\alpha=2$, i.e., interference with Gaussian distribution.)
% Based on the convergence framework with $\alpha$-norm, theoretical results demonstrate that interference with heavier tail (i.e., larger $\alpha$) would lead to a slower convergence rate for the analog \ota{} edge learning algorithms, which is identical to the findings of empirical results. 
% Moreover, scaling up the \ota{} edge learning system by increasing the total number of UEs would effectively mitigate the influence of the interference and speed up the convergence from both theoretical and empirical perspective.

In summary, the bad face of the heavy-tailed interference in analog \ota{} computations brings challenges as well as the opportunity for both theoretical analysis and algorithm design. A general framework could be developed to tackle the performance degradation by capturing the intrinsic property of the heavy-tailed interference~\cite{zhang2022turningpca,YanCheQue:21JSTSP}.

\section{The Positive Side of Interference}  \label{sec:good_face}
As we know, nothing interesting is ever completely one-sided, so as interference. 
In this section, the  \textit{good face} of interference, i.e., its potential benefits on the generalization capability, convergence acceleration, and privacy protection in analog \ota{} FL will be discussed.

To commence with, as an epitome among the diverse ML techniques, deep learning (DL) achieves great success in various areas and  applications. 
In practice, stochastic gradient descent (SGD) or mini-batch SGD is one of the de facto choices for training DL models.
Concurrently, the gradient noise would be introduced due to randomness in the SGD algorithm. 
Recent work reveals that the gradient noise in SGD tends to exhibit a heavy-tailed behavior with highly non-Gaussian characteristics  \cite{simsekli2019tailindex}.

On the other hand, both the empirical and theoretical findings demonstrate that the gradient noise in SGD can guide the optimizer to find a flat minimum that possesses a good generalization power. 
In consequence, perturbed gradient descent (PGD) methods have been investigated by several works,  demonstrating that injecting gradient noise during the training process can effectively improve the generalization performance as well as avoid over-fitting \cite{orvieto2022anticorrelated}. 
The reason mainly attributes to that artificial noise can help the model escape local minimum and saddle points. 

Following our discussions in the previous sections, since the wireless interference often obeys a  heavy-tailed distribution, it could be utilized as a natural perturbation to the locally updated gradients from UEs in such GD-based analog \ota{} edge learning.
Another question then begs:  ``\textit{Can the channel noise play a similar role as the artificial gradient noise to boost the FL performance?}''
% The following part would demonstrate the good face by addressing 

\subsection{Potential Generalization Enhancements} \label{subsec:better_generalization}
Taking the induced interference as a natural perturbation  to the aggregated gradients in \ota{} edge learning, it is worthwhile to explore the relationship with the generalization capability. 
 
Interestingly, recent work established theoretical bounds to show that it is possible to achieve a  better generalization in the presence of heavy-tailed interference \cite{YanCheQue:21JSTSP}. 
More precisely, in the analog \ota{} edge learning system, the generalization error would decrease  along with the interference tail index.  
Even though this finding may be empirically counter-intuitive with regard to the conventional analysis of the interference, it would be a promising application by developing approaches to reap this potential gain.
\begin{figure}[t!]
  \centering
  \subfigure[\label{fig:5a}]{\includegraphics[width=0.95\columnwidth]{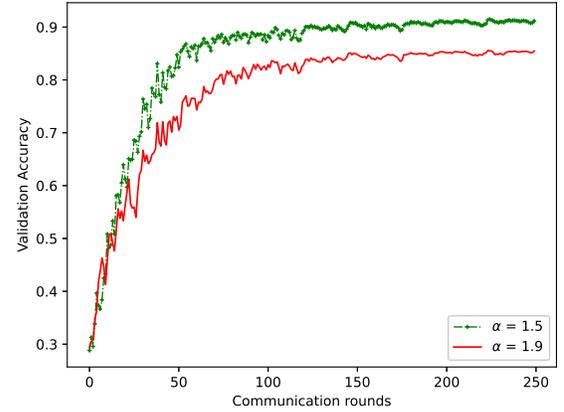}} ~
  \subfigure[\label{fig:5b}]{\includegraphics[width=0.95\columnwidth]{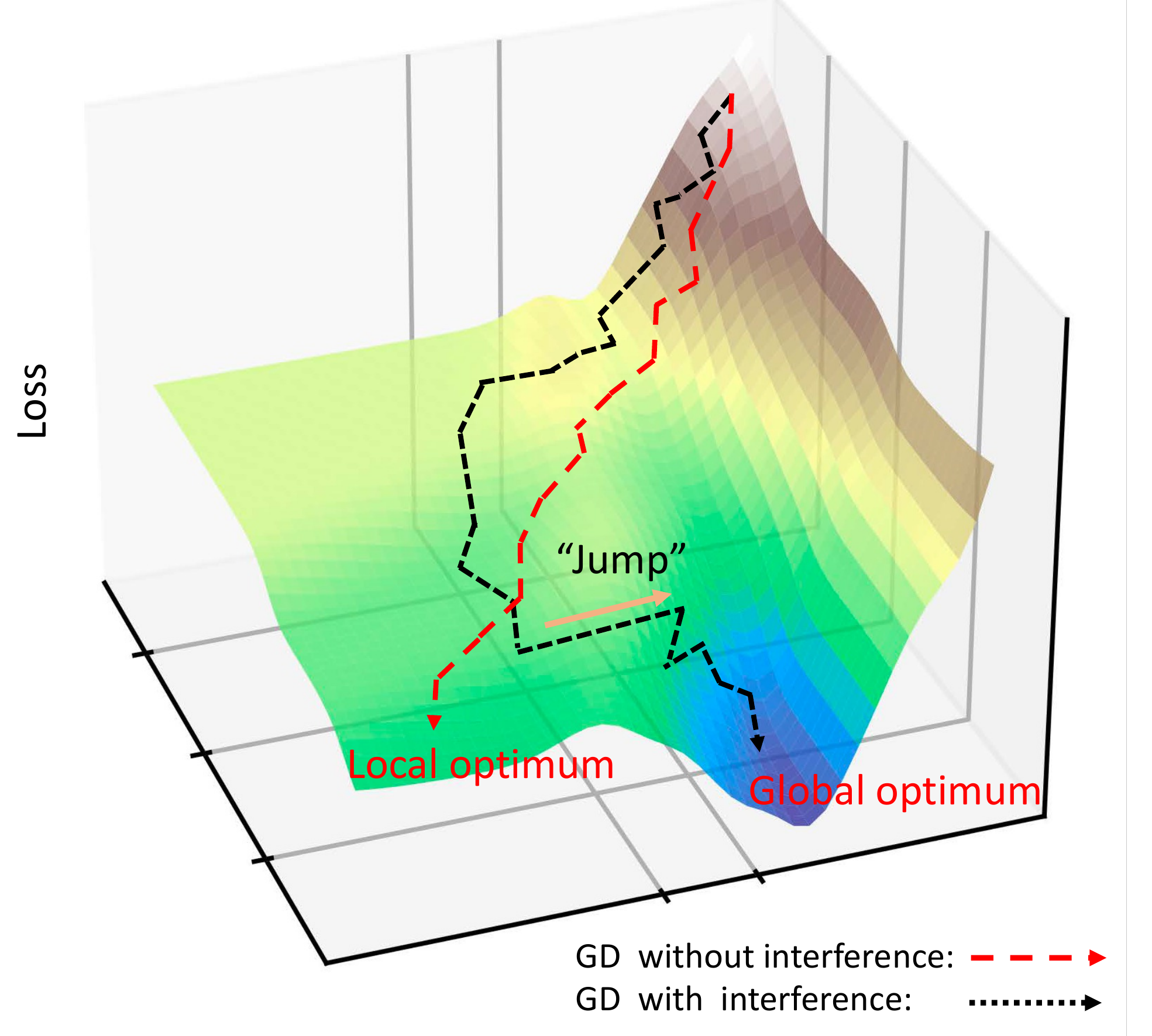}}
  % \subfigure[\label{fig:5b}]{\includegraphics[width=0.95\columnwidth]{Figure/landscape.pdf}}
  \caption{ A depiction of good face of $\alpha$-stable interference: ($a$) gives simulation results of the training loss of MLP on the MNIST data set, with different tail index $\alpha$ and ($b$) gives an illustrative example via a loss landscape that interference helps optimizer jump out of local minimum. }
  \label{fig:alpha_generalization}
\end{figure}

As shown in Fig. \ref{fig:5a}, the interference in analog \ota{} computations with a smaller value of tail index $\alpha$ (i.e., a heavier tail in the distribution) could achieve better generalization performance with respect to the test accuracy of the trained ML model. 
This benefit is ascribed to the possible occurrence of the strong impulse of the heavy-tailed interference. 
For better illustration, we provide a loss landscape with the task of training a 3-layer neural network on MNIST dataset.
As exemplified by Fig.~\ref{fig:5b}, when the optimizer is trapped in the local optimal status, such a ``big jump'' could help the global model escape from the local minima.  

It is also important to stress that this phenomenon occurs with a certain probability because the change of the channel status is independent of the status of model training.
Therefore, it would be another open question as well as the potential research opportunity to develop a method or framework to utilize the good face of the interference and further enhance the performance for analog \ota{} edge learning systems.

\subsection{Accelerating Model Training with Power Control}
% As mentioned in Sec. \ref{subsec:better_generalization}, the random noise in wireless channel has the potential to 
The previous part shows that interference may help the optimizer jump out of the local optimum or escape from the saddle point during the GD-based training process.
However, the mentioned phenomenon of generalization enhancements only occurs with a certain probability due to the randomness  of channel interference. 
Then, a natural consideration is to maneuver the interference when the optimizer falls in the region of local optimum and/or saddle points, so as to assuredly reap the performance gain in  analog \ota{} edge learning.

Motivated by the above issue, a practical approach is proposed to harness the interference in analog \ota{} computations to accelerate training of distributed principle component analysis (PCA) \cite{zhang2022turningpca}.
The acceleration is based on the detection of saddle points. Once a saddle point is detected during training, an online power control mechanism in network management would be applied to amplify the effect of noise to help the GD-based optimizer escape from the saddle points and realize acceleration.

Therefore, via a proper manipulation on the noisy effect based on the real-time training status, the interference could indeed be repurposed to play a positive role in the training process and accelerate the convergence of analog \ota{} edge learning system.

\subsection{Privacy Preservation and Efficient Sampling}
Differential privacy (DP) is a widely adopted standard to quantify the privacy leakage of sensitive data, which can be generally achieved by randomly perturbing the data.
In the context of analog over-the-air federated edge learning, random interference serves as a natural perturbation to the uncoded signal, thereby enhancing the privacy protection of the end-users' dataset. 
Particularly, considering that channel noise has different influences on the convergence and privacy aspects, power control can be used to actively keep the signal-to-noise ratio (SNR) at an acceptable level while ensuring the DP constraint\footnote{The level of DP typically depends on the degree of perturbation imposed by the randomized mechanism. The DP performance in this case would be affected by the SNR adjusted via power control.}. In fact, with an adequately devised power control mechanism, privacy can be obtained for free under certain DP constraints with guaranteed learning performance~\cite{liu2020privacyfree}.
% \textcolor{red}{Differential privacy (DP) is a widely adopted standard to quantify the privacy leakage of sensitive data, which could be achieved via adding random perturbation to the data.}
% As a natural perturbation to the uncoded analog signal, the random interference in wireless networks could serve as an enhancement technique to the privacy protection in analog \ota{} edge learning. 
% \textcolor{red}{Particularly, considering that channel noise has different influence on the convergence and privacy level, power control could actively keep the signal-to-noise ratio (SNR) in a sufficient low level while ensuring the DP constraint, assisted with obtained channel state information (CSI).} 
% With adaptive power control, such privacy could be obtained for free with DP constraints \textcolor{red}{with learning performance guarantee}\cite{liu2020privacyfree}. 

In the context of Bayesian federated learning, especially the gradient-based Markov Chain Monte Carlo (MCMC) method, randomness also exists in the MCMC sampling.
Then, the interference would hold a double role of the randomness for approximation in MCMC and privacy protection in such a scenario.
Both theoretical and empirical results reveal that, the channel interference could be repurposed under a suitable condition to make the system with noisy communications achieve the same performance as with the ideal communication  \cite{liu2021wirelessmcmc}.

In summary, the inherent channel interference in analog \ota{} computations could also bring benefit to the training and privacy performance of federated edge learning. Considering both the bad face and good face  of the interference in analog \ota{} edge  learning system, this two-sided effect would provide a new perspective to designing integrated communication and computation framework for robust, communication-efficient, and privacy-preserving edge learning.

\section{Future Trends and Open Issues}
The above discussions have amply demonstrated interference's effects on federated edge learning: On the positive side, it has the potential of achieving ``one stone, many (and many) birds"; on the negative side, it could be ``the straw that broke the camel's back", whereas a very strong electromagnetic impulse may damage the whole trained results.
To harvest the potential gains and cope with the existing crux, we summarize the following directions for future research pursuits:
% By exploiting the analog over-the-air computing with the two-sided effects of the random channel interference, we discussed both the potential benefits and the issues within the paradigm of the federated edge learning. 
% Both the bad face and good interference could be addressed to further empower the future intelligent wireless network with the analog \ota{} computations.
% On the one hand, tackling the negative impacts of interference with new optimization and control mechanism would accelerate training with improved stability performance.
% On the other hand, capturing the good face via repurposing interference could obtain performance gain with privacy protection.
% An integrated design by considering both two faces of the interference shall be explored. 
% A few potential research opportunity is listed as follows:
% The future trends of \ota{}-based edge learning shall further explore the advances in the aspects of both the analog transmissions and distributed learning.
% Concurrently, there still remains several open issues for further exploring this two-face property and then enhancing the performance of edge learning system.
\begin{itemize}
    \item \textit{How to conduct pruning and adaptive optimization to improve system efficiency?} Most of the current analog \ota{} methods have no requirements or assumptions on the optimizations and training process in both the edge server and local sides. Consequently, some communication-efficient distributed learning approaches (e.g., pruning, quantization, and masking) and federated optimization methods can be further incorporated into the \ota{}-based federated edge learning systems. 
    Applying the pruning or quantization method at the side of UEs for the model weight in the uplink transmission, better communication efficiency could be achieved in the bandwidth-limited network. 
    How to mitigate the variance and fluctuations brought by the interference in \ota{} computations could be further investigated in these cases. 
    
    \item \textit{How to stabilize the training process?} With the existence of the heavy-tailed interference in the analog \ota{} edge learning system, a better control mechanism at the edge server is of necessity to smooth out the fluctuations and hence improve robustness in the training process. Adaptive optimization and robust training methods are possible candidate techniques for this purpose. For example, federated optimization methods, e.g., FedAvgM and FedOpt with different adaptive optimizers, can be applied on either the (global) edge server-side or (local) UE side to enhance the stability performance. Furthermore, robust training techniques like \textit{gradient clipping} could also be used during local training to eliminate the excessive deviations in model parameters and thus stabilize the training performance. 
    
    \item \textit{How to re-purpose interference for better generalization?} The mentioned potential gain on generalization performance relies on the random noise effect to jump out of the local minimum. It remains an open question about the development of methods to repurpose the interference for generalization capability enhancements. Following experience in the previous trials, one needs to jointly consider the local geometric landscape of the optimizer and the instantaneous variations in the interference. On the other hand, more advanced wireless technologies such as multiple antenna arrays and intelligent reflection surfaces can be employed to not just enhance the received signal quality, but, more importantly, engineer the interference distribution for better performance of the edge learning system.

    \item \textit{How about temporal correlation in the interference noise?} In practice, interference does not always vary over time in an i.i.d. manner. 
    Actually, the correlations of interferers' spatial locations and fading effect among a transmission area would lead to temporal correlation in the interference noise. 
    And the effect of such a temporally correlated interference on the performance of analog \ota{} edge learning system is another open question.
    
    \item  \textit{What about data corruption in the end-user equipments?} Considering a more realistic edge learning system, the diversity among different UEs' hardware reliability, bias, and preference should be taken into account in the design of edge learning systems and network management. In such a system, we shall get rid of the ideal assumption that all the local training data and labels are correct. In particular, we may consider heterogeneous label noise exists among UEs, where a subset of data may have incorrect labels. It is necessary to design a robust federated edge learning system for network management in the presence of  both channel interference and local label noise. And whether the potential benefits of the interference exist or not for corrupted local data remains as an open question. 
    
    \item \textit{How to personalize analog \ota{} FL system?} \textit{Personalizing} the FL model has become a trending approach to tackle the poor generalization issue of conventional FL framework that only focuses on a single generic global model and hence not adapting well to the diverse local data distribution in heterogeneous wireless networks. Recently, it was proposed to utilize personalized model training to realize the enhancements on robustness and fairness of FL systems. As the induced channel interference in analog \ota{} computations 
   and the distributed real-world noisy training data would possibly harm the training performance, the personalized  local training procedures would have the potential to be further explored in the analog \ota{} edge learning system to enhance the robustness and stability to mitigate the performance fluctuations in such scenarios. For instance, each client could maintain its own personalized local training objectives or model architectures.

   \item \textit{How to design a secure analog over-the-air computing framework with diverse attacks?} Even though the interference exhibited enhanced DP protection to the transmitted data, it remains unclear whether the induced interference could help defend against the gradient-based attacks, especially for neural network-based tasks, such as gradient inversion. % and the data distribution inference based on the model weight of local model classifiers. 
   Another promising topic is to explore the backdoor attack and the corresponding defense in analog over-the-air computations. It is particularly worthwhile to investigate the possibilities of conducting proactive adversarial perturbation or defense to combat backdoor attacks via harnessing the induced interference.
   In addition, how to defend against signal interception remains to be further explored.

\end{itemize}

\section{Concluding Remarks}
By exploiting the superposition property of wireless waveforms, analog \ota{} computations provide a high-scalability and concurrently low-cost solution to federated edge learning systems. In this article, the two-sided effect of the interference in the analog \ota{} edge learning system has been investigated. Specifically, while interference impedes the convergence rate and threatens system stability, there are also silver linings of this factor, that it has the potential to enhance end-user privacy, help the training model escape saddle points or local minimum, and attain a better generalization. Such an effect appeals to designing new optimization frameworks as well as control mechanisms in intelligent network management to repurpose the channel interference in analog \ota{} computations to enhance the performance \ota{} edge learning.

%\balance
\bibliographystyle{IEEEtran}
\bibliography{bib/StringDefinitions,bib/IEEEabrv,bib/edge_ota}

\end{document}